\documentclass[journal]{journal}
\usepackage{cite}
\usepackage[none]{hyphenat}

\usepackage[justification=centering]{caption}	

\usepackage[labelsep=space]{caption}		

\usepackage[font=footnotesize]{caption}
\renewcommand{\figurename}{Fig.}

\ifCLASSINFOpdf
\usepackage[pdftex]{graphicx}
\graphicspath{{../images}{../jpeg/}}
\DeclareGraphicsExtensions{.pdf,.jpeg,.png}
\else
\fi
\usepackage[cmex10]{amsmath}
\usepackage{array}
\usepackage{mdwmath}
\usepackage{mdwtab}
\usepackage{eqparbox}
\usepackage[tight,footnotesize]{subfigure}
\usepackage{caption}
\usepackage[font=footnotesize]{subfig}
\usepackage{fixltx2e}
\usepackage{stfloats}
\usepackage{url}
\usepackage[english]{babel}
\usepackage[utf8]{inputenc}

\usepackage[nottoc]{tocbibind}

\usepackage{nomencl}
\makenomenclature

\begin{document}
\title{Quantum-Assisted Feature Selection for Vehicle Price Prediction Modeling
%\author{David Von Dollen, Florian Neukart, Thomas Bäck, Megan Michaels, Daniel Weimer

%\author{
%\IEEEauthorblockN{David Von %Dollen\IEEEauthorrefmark{1}\IEEEauthorrefmark{3}, Florian %Neukart\IEEEauthorrefmark{2}\IEEEauthorrefmark{3}, 
%Daniel Weimer\IEEEauthorrefmark{1},
%Thomas Bäck\IEEEauthorrefmark{3}}
%\thanks{\IEEEauthorblockA{\IEEEauthorrefmark{1}Volkswagen %Group of America, Auburn Hills MI, U.S.A}\\
%\IEEEauthorblockA{\IEEEauthorrefmark{2}Volkswagen Group %Data:Lab, Munich Germany}\\
%\IEEEauthorblockA{\IEEEauthorrefmark{3}LIACS, Leiden, %Netherlands\\
%Email: \IEEEauthorrefmark{1}\IEEEauthorrefmark{3}david.vondollen@audi.com
%   }}
%}}

\author{
\IEEEauthorblockN{{David Von Dollen, Florian Neukart, Daniel Weimer, Thomas Bäck}
}
}

\thanks{\IEEEauthorblockA{David Von Dollen and Daniel Weimer are with Volkswagen Group of America, Auburn Hills MI, U.S.A}\\
David Von Dollen and Florian Neukart are with Volkswagen Group Data:Lab, Munich Germany\\
David Von Dollen, Florian Neukart, and Thomas Bäck are with  LIACS, Leiden, Netherlands\\
Corresponding Author Email:david.vondollen@audi.com}
%  }}
%}}
}
\maketitle

\thispagestyle{empty}

\begin{abstract}
%\boldmath
Within machine learning model evaluation regimes, feature selection is a technique used to reduce model complexity and improve model performance in regards to generalization, model fit, and accuracy of prediction. However, the search over the space of features to find the subset of $k$ optimal features is a known NP-Hard problem. In this work, we study metrics for encoding the combinatorial search as a binary quadratic model, such as Generalized Mean Information Coefficient and Pearson Correlation Coefficient in application to the underlying regression problem of price prediction. We investigate trade-offs in the form of run-times and model performance, of leveraging quantum-assisted vs. classical subroutines for the combinatorial search, using minimum redundancy maximal relevancy as the heuristic for our approach. We achieve accuracy scores of 0.9 (in the range of [0,1]) for finding optimal subsets on synthetic data using a new metric which we define. We test and cross validate predictive models on a real world problem of price prediction, and show a performance improvement of mean absolute error scores for our quantum-assisted method $(1471.02 \pm{135.6})$,  vs. similar methodologies such as recursive feature elimination $(1678.3 \pm{143.7})$. Our findings show that by leveraging quantum assisted routines we find solutions which increase the quality of predictive model output while reducing the input dimensionality to the learning algorithm on synthetic and real-world data. 
\end{abstract}


% \fi
\IEEEpeerreviewmaketitle

\section{Introduction}
\IEEEPARstart{D}{ata} pre-processing techniques, along with exploratory analysis and feature engineering, are standard steps within pipelines for predictive model training and deployment \cite{Kotsiantis2007DataPF}. In practice, before feeding data to a learning algorithm, it is common to cleanse data sets in order to improve predictive performance and generalization capability of models. 

Feature selection (FS) is a pre-processing technique, in which a subset of features are selected from a data set which may contain more relevant information than the total set of features, due to co-linearity, redundant, or constant features within the total data set. Benefits to utilizing FS within a pre-processing step include reduced data dimensionality, better scaling for learning algorithms, and reduced costs in regards to model training and fitting. Feature selection differs from other techniques such as Feature Engineering in that it does not create a projection of linear combinations of features into a new subspace, or a change of basis using an eigendecomposition of a covariance data matrix, as in the case of Principal Components Analysis \cite{pearson_karl_1901_1430636}, but rather filters down and eliminates features with low information in regards to a target variable. This is an important property, since the preservation of the original feature space allows for higher visibility and explainability when accounting for predictive model output. Feature Selection techniques may leverage an output from an estimator or model in the selection process, also known as $wrapper$ methods \cite{10.5555/645527.657288} or may use heuristics and select an optimal subset, also know as $filter$ methods. In this work we examine using quantum-assisted methods to find a combinations of features of a subset of size $k$, which may show improved performance characteristics and a decrease in the input feature space for our learning algorithms, which is a known NP-Hard combinatorial optimization problem \cite{1997TheMF}. 

Our contribution is as follows: framing the combinatorial search as a binary quadratic model, we examine leveraging a quantum device to search the combinatorial space of finding optimal feature subsets of size $k$. We investigate various distance and correlation metrics for formulation of the binary quadratic optimization problem. We apply the heuristic of Maximal  Relevancy Minimal  Redundancy (MRMR) \cite{1453511} with the idea that we want to find a subset of features which may have strong predictive signal in regards to a target variable (Maximal Relevancy), but low pairwise feature correlation between independent variables (Minimal Redundancy). We train and compare two types of regression models using quantum assisted feature selection along with benchmark selection methods over all features, greedy selection, and a wrapper method, over two data sets with continuous target values.  We examine run times and model predictive performance  using this methodology, and apply to a real-world problem of data pre-processing for predictive models for vehicle prices. We show that by using quantum-assisted routines, we find combinations of features which increase the predictive quality of models on validation sets of data, and improve upon our benchmarks of all features, greedy feature selection, and recursive feature elimination.
\section{Related Work}
The problem of Feature selection is well-studied in the literature \cite{BLUM1997245,Motoda2002FeatureSE}. Recent approaches apply mutual-information based metrics for feature selection for supervised learning in regards to classification applications \cite{10.1145/2623330.2623611}. In recent years, new correlation metrics have been proposed which may have more expressive power in measuring relationships between variables. Examples of this are Maximal Information Coefficient (MIC) \cite{Reshef2013EquitabilityAO} and Generalized Mean Information Coefficient (GMIC) \cite{Luedtke2013TheGM}, which introduce the concept of $equability$ in variable relationships, and may be more robust when dealing with non-linear relationships than correlation statistics such as Pearson correlation coefficient, which assume linear relationships.

In recent years, the availability of quantum devices leveraging quantum processing units (QPUs) have come online, and applications have been developed leveraging these devices for solving real world problems within various industries. For example, in \cite{10.3389/fict.2017.00029}, the authors leveraged a quantum annealing system to optimize traffic flows around the city of Beijing, and the authors in \cite{Stamatopoulos2020optionpricingusing} showed how to price options using quantum algorithms run on a gate-model quantum chip. 

The feature subset selection problem was formulated by the authors in \cite{JMLR:v11:rodriguez-lujan10a} as a quadratic program, where mutual information and Pearson correlation coefficient were used to calculate the matrix $\mathbf{Q}$, or a symmetric positive semi-definite matrix representing quadratic terms for minimizing the multi-variate objective function. There have been research efforts to apply quantum annealing to searching feature space for optimal subsets using mutual information based formulations of interactions and linear terms for Ising spin-glass models and quadratic unconstrained binary optimization \cite{Rounds2017OptimalFS,sharma2019quantum}. Some of these efforts have included complexity considerations in regards to scaling for the quantum-assisted feature subset selection problem, claiming performance gains of $O(1/m^2)$ versus $O(mn^2)$ for classical computation \cite{sharma2019quantum}. This work in particular, claims a bound for the quantum assisted routine given by the size of the minimum gap $g(t) = E0-E1$ in the energy eigenvalues during in the annealing regime, with a resulting time complexity of $T = O(\frac{1}{g^2min})$ where $T$ is the upper time limit.

In regards to the price prediction problem, this has been well studied in the literature for machine learning research \cite{ Pudaruth2006PredictingTP,Monburinon2018PredictionOP, 10.5120/ijca2017914373}. In \cite{Pudaruth2006PredictingTP}, the authors built predictive models for price prediction for used cars in Mauritius. The models included naive bayes and decision tree algorithms, which contained a classification step with reported accuracy rates in a range of 60-70\%, and achieved a mean error of 51000 and 27000 for the regression component using linear regression. Monburinon et. al tested various regression models for price prediction for German used cars in \cite{Monburinon2018PredictionOP}, with gradient boosted decision trees outperforming random forest and multiple regression (mean squared error = 0.28). In \cite{10.5120/ijca2017914373} the authors applied feature selection techniques for multiple regression models for price prediction and reported score in regards to model fitness ($R^2$ = 0.9861).

 
%\hfill January 11, 2007

\section{Methods}

\subsection{The Feature Subset Selection Problem}
Consider a data set:
\[\mathcal{D} = \{\mathbf{X}\in \rm I\!R^{N\times M} , \mathbf{y}\in \rm I\!R ^N\} \]
where $\mathbf{X}$ is a data matrix of size $N \times M$ of $N$ rows, and $M$ columns,
and $\mathbf{y}$ is a column vector of size $N \times 1$ of $N$ rows, and $1$ column. We wish to find an approximate functional mapping or hypothesis of $h(\mathbf{X}) \simeq \mathbf{y}$ using various learning algorithms. This is also known as supervised learning, and our goal is to minimize a generalization error for a given loss function between predictions and a target variable in order to make predictions given new data, or find the best hypothesis from the hypothesis space which the given learning algorithm encompasses. 

In the feature subset selection problem we wish to find a subset of features $\mathbf{X_k}$ where each row vector $\mathbf{x_i}$ in $\mathbf{X}$ (where $i =  \{1, 2,... N\}$) is of reduced column size $k$, i.e each row vector is filtered down as in \{$x_{i1}, x_{i2}, ..x_{ik}\}$ where $k < M$. We may investigate various values of $k$ such that the loss for our function $h(\mathbf{X_k}) \simeq \mathbf{y}$ is less than or equal to $h(\mathbf{X}) \simeq \mathbf{y}$ cross validated on a hold out test set of data. Essentially the feature subset selection problem is one of choosing the optimal subset of features or columns from $\mathbf{X}$ of size $k$. In this problem, we assume that there exists an optimal subset of size $k$, which may not be the case for all data sets, where the optimal set $h(\mathbf{X_k}) = h(\mathbf{X})$ or where $k = M$.

\subsection{Relevancy, Correlation and Distance Metrics}
In order to compare features with a target variable, and investigate pair-wise relationships amongst the feature set $\mathbf{X}$, we must first define our distance functions with which we formulate the binary optimization problem for the feature subset selection. In our experiments we will switch out and compare each distance function when used to model linear and quadratic terms for a binary quadratic model. 

We examined Maximal Information Coefficient (MIC), Generalized Mean Information Coefficient (GMIC), Mutual Information (MI), and Pearson Correlation Coefficient (PCC) for this study. Let us define each in the following:

Mutual Information (MI) is defined as:
\begin{equation}
MI(\mathbf{x, y}) = \sum_{\mathbf{x, y}} P(\mathbf{x, y}) \ln{{P(\mathbf{x, y})}\over{P(\mathbf{x}) P(\mathbf{y})}}   
\end{equation}

In our case we are looking at mixes of purely continuous as well as continuous and discrete variables, with which there are various strategies for binning continuous values to estimate probabilities for the MI calculation. These include kernel density estimation, binning continuous to discrete variables, and clustering algorithms. For our method we use the $k$-nn binning strategy from \cite{10.1371/journal.pone.0087357} to estimate the MI:
\begin{equation}
I(\mathbf{x, y}) = \psi{(N)}- \langle\psi{(\mathbf{x})}\rangle +\psi{(\kappa)} - \langle\psi{(\mathbf{m})}\rangle
\end{equation}
Where $\psi$ is the digamma function, $\kappa$ is the distance for the designated nearest neighbor and $m$ is the count of number of neighbors. For more details please also see \cite{doi:10.1080/01966324.2003.10737616}.

Various algorithms have been proposed to handle a mix of discrete and continuous and purely continuous algorithms based on mapping mutual information to a grid. MIC and GMIC belong in this category.

MIC is defined as \cite{Luedtke2013TheGM}:

First we define a $Characteristic$ Matrix $C$:

\begin{equation}
C(\mathbf{x, y})_{i,j} = \frac{I^*(\mathbf{x,y})_{i,j}}{\log_2 \min\{i, j\}}
\end{equation}
Where $I^*(\mathbf{x,y}$ represents the binned values of $\mathbf{x,y}$ in a grid $G_{ij}$.
Then we obtain the MIC using this characteristic matrix:
\begin{equation}
MIC(\mathbf{x,y}) = \max_{ij<B(n)}\{C(\mathbf{x, y})_{i,j}\}
\end{equation}

where $B(n) = n^{0.6}$ is a maximal grid size as recommended in the original text \cite{Reshef2013EquitabilityAO}. Note that we consider $ij$ in this equation as a product notating grid size.

We can define GMIC by taking the characteristic matrix and extending it to find the maximal characteristic matrix:
\begin{equation}
C^*(\mathbf{x, y})_{i,j} = \max_{kl<ij)}\{C(\mathbf{x, y})_{kl}\}
\end{equation}
Where we again use the terms $kl$ and $ij$ to denote grid sizes. We may then use this to define the GMIC measure as defined in \cite{Luedtke2013TheGM}. 

\begin{equation}
 GMIC(\mathbf{x, y}) = \left( \frac{1}{Z} \sum_{ij<B(n)}(C^*(\mathbf{x, y})_{i,j}^p \right) ^{1/p}
\end{equation}

Where $p \in [-\infty, \infty]$. For our work we set $p = -1$ and $Z$ is the primorial of $(i,j)$ where $ij \leq B(n)$ as outlined in the original work \cite{Luedtke2013TheGM}.

There exists a lively debate in the literature as to the significance of the statistical power of these methodologies \cite{Kinney201309933}. We note that we incur additional overhead in regards to computational cost in allocating grids, as well as tuning parameters such as $B(n)$ in the case of MIC and $p$ in GMIC using these methods.

In our study we reviewed the performance of Pearson Correlation Coefficient (PCC). This statistical measure is ubiquitous in science and engineering for measuring linear relationships between variables.

Pearson correlation coefficient is defined as \cite{math_vault_2020}:
\begin{equation}
PCC(\mathbf{x, y}) = \frac{ \sum_{i=1}^{n}(x_i-\bar{x})(y_i-\bar{y})}{\sqrt{\sum_{i=1}^{n}(x_i-\bar{x})^2}\sqrt{\sum_{i=1}^{n}(y_i-\bar{y})^2}}
\end{equation}
Where $\bar{x}$ and $\bar{y}$ are the mean of $\mathbf{x}$ and $\mathbf{y}$.

%Similar to PCC, distance correlation (DC) is used to look at relationships between variables, however it has been claimed to be more robust to non-linear relationships than PCC.
%Distance correlation is defined as \cite{2020SciPy-NMeth}:
%\begin{equation}
%DC(\mathbf{x,y}) = 1 - \frac{(\mathbf{x}-\mathbf{\bar{x}}) \cdot (\mathbf{y}-\mathbf{\bar{y}})}{\left\Vert(\mathbf{x}-\mathbf{\bar{x}})\right\Vert_2 \left\Vert(\mathbf{y}-\mathbf{\bar{y}})\right\Vert_2}
%\end{equation}

\subsection{QUBO Formulation}
In order to formulate our problem as a quadratic binary optimization model (QUBO) we first need to calculate a distance measure for each column vector $i$ in $\mathbf{X}$ vs. $\mathbf{y}$. In this case, we use the absolute value of the distance measure and negate it as we wish to find a minimum for our optimization problem. These values then become the linear terms along the diagonal of the $\mathbf{Q}$ matrix.

With these linear terms, we encode the $maximal \ relevancy$ portion of our heuristic, although we negate the values as the optimization takes the form of finding the minimum of the domain. By taking the absolute value of the distance function, we give greater weight to features that have higher relevancy  or correlation with the target variable by treating positive and negative correlation equally.

Then, we calculate the distance between each pairwise column vectors indexed at $i$ and $j$ in $\mathbf{X}$. This allows us to formulate the interactions for the quadratic terms for the binary quadratic model, which become values along the upper-diagonal of the $\mathbf{Q}$ matrix. This encodes the $minimum \ redundancy$ characteristic of our heuristic. We want to find combinations of features that are not correlated, or are more distant from each other, while retaining relevancy to the target variable.
\begin{equation}
\mathbf{Q}_{ij} = 
    \begin{cases}
    - |distance(\mathbf{X}_i, \mathbf{y})|, & \text{if}\ i=j \\
    |distance(\mathbf{X}_i, \mathbf{X}_j)|, & \text{if}\ i < j \\
    0,  \text{otherwise}
    \end{cases}
\end{equation}

We combine these to obtain our QUBO formulation for our optimization problem. For sake of clarification, we use the $\mathbf{\omega}$ term here to represent a vector of qubit values. We include a scaling parameter $\alpha$ that we use to scale the domain of the optimization landscape.
\begin{equation}
E(\mathbf{\mathbf{\omega}}) = \alpha\sum_{i\leq j}\mathbf{\mathbf{\omega}}_i\mathbf{Q}_{ij}\mathbf{\omega}_j \qquad \mathbf{\mathbf{\omega}} \in \{0, 1\}
\end{equation}

Finally, we impose a penalty term, such that the resulting sample from our objective function only has $k$ qubits turned "on", or have a value of 1. We  introduce a scaling parameter for the penalty term, $\lambda$, to enforce this constraint.

\begin{equation}
E(\mathbf{\mathbf{\omega}}) = \alpha\sum_{i\leq j}\mathbf{\mathbf{\omega}}_i\mathbf{Q}_{ij}\mathbf{\omega}_j  + \lambda(\sum_i\mathbf{\omega}_i - k)^2
\end{equation}

We then use this formulation to follow an annealing schedule and sample solutions to find a minimum energy $E$. In our resulting vector $\mathbf{\mathbf{\omega}}$, which is of $M$ length, our constraint ensures that $k$ qubits have a value of 1, and the rest 0. We then use this vector to filter the data set $\mathbf{X}$ to obtain $\mathbf{X_k}$, where the $k$ columns are filtered from the index location where $\mathbf{\omega}_m$ = 1 for \{ $\omega_1$, $\omega_2$, \dots ,$\omega_m$\}.

%\IEEEpubidadjcol

\section{Experiments}
\subsection{Data Sets}
For our experiments, we tested our feature selection algorithms on two data sets, one synthetic, and one drawn from real world samples.

\subsubsection{Friedman 1}
We generated data as specified in \cite{friedman1991} . Here, the data is generated by the function:
\begin{equation}
    \mathbf{y} = 10  \sin(\pi x_1x_2) + 20(x_3 - 0.5)^2 + 10x_4 + 5x_5 + \epsilon
\end{equation}
for each row $x_i \in \mathbf{X}$ where $\epsilon$ is some random, normally distributed noise, and the rest of the features are independent and drawn from a uniform distribution on the interval of [0,1]. We generated 100 instances of this data for our train/test 70/30 percentage split, with a feature size of 50. We specifically tested on this data set in order to study the difference in performance using various mutual information based metrics within the QUBO formulation of the optimization problem, since the generating function contains a non-linear first term, and one of the reported gains in using these distance metrics is in measuring non-linear relationships. Another advantage is experimenting with this data set is that the optimal subset of features are known in advance, and therefore we may determine how accurate the feature selection algorithms are in response to the optimum.

\subsubsection{UCI Automobile data}
The next data set that we tested on was for vehicle price prediction using the open source Auto data set from the UCI machine learning repository \cite{Dua:2019}. In this data set, we have prices for 205 automobiles, along with other features such as fuel type, engine type, and engine size. We encoded all categorical and nominal features using ordinal encoding, which preserved the attribute size of 26.

\begin{figure}[htp]
    \includegraphics[width=\linewidth]{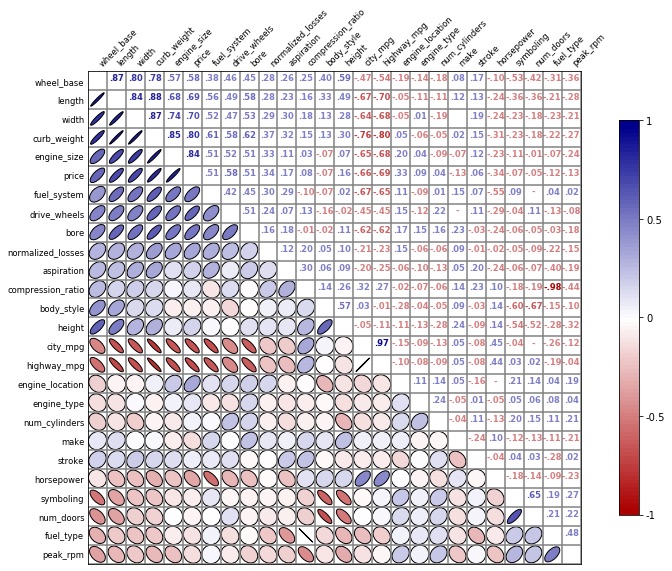}
    \captionsetup{font=footnotesize}
    \caption{Correlation Plot for UCI Auto data $\mathcal{D} = \{\mathbf{X}, \mathbf{y}\}$, using Pearson Correlation Coefficient as a distance metric between features. Higher values of positive correlation are indicated in dark blue, and negative correlation in red. In the feature subset selection problem, we wish to select a subset of features such that the relevancy is maximized with regards to a target variable, in this case, price, and the feature to feature correlation amongst independent variables is minimized.}
    \label{Correlation for UCI Auto Data}
\end{figure}

%\begin{figure}[htp]
%    \includegraphics[width=\linewidth]{images/corr_plt2.png}
%    \caption{Correlation Plot for sample of filtered UCI Auto data, $\mathcal{D} = \{\mathbf{X_k}, \mathbf{y}\}$, using Quantum Assisted method with Pearson Correlation Coefficient as a distance metric between features, where $k$ = 6}
%    \label{Correlation for UCI Auto Data}
%\end{figure}

In performing an exploratory analysis of the Auto data, we examined the correlation between features and target variable. In reviewing Fig. 1 for the UCI Auto data, we notice that there are strongly correlated features with the target value of price. In terms of positive correlation, we see curb weight (PCC= 0.80) and engine size (PCC= 0.84) as having a strong linear relationship, and city and highway miles per gallon (PCC = -0.66, PCC= -0.69)  as having a strong negative correlation.

%\subsubsection{VW Auto pricing data}
%We also tested on a real-world data set of auto pricing data with a sample of 100 instances. The data was anonymized to preserve customer privacy, and included 48 attributes such as number of cylinders and doors.

\subsection{Estimators}
\subsubsection{Predictive models}

During the training/testing regime we used a 70/30 percentage split between training and test sets. We measured performance of two different types of predictive models for the underlying regression problem. These models included the following:

Multiple linear regression (LR):
We used a multiple linear regression model to estimate predicted values of $\mathbf{\hat{y}}$ \cite{Free:2005}.
Multiple Linear Regression is defined as:
\begin{equation}
    \mathbf{y} = \mathbf{X} \mathbf{w} + \mathbf{\epsilon}
\end{equation}
Where $\mathbf{w}$ are parameters which we wish to optimize $\mathbf{w}$ such that $\epsilon = \mathbf{y} - \mathbf{X}\mathbf{w}$ is minimized. We may then substitute new test data $\mathbf{X}$ in with our trained parameters to obtain $\mathbf{\hat{y}}$. We use the term Multiple linear regression to clarify that we have two or more independent variables for which we would like to find a mapping to a target variable.

Gradient boosted regression trees (GBR):
We looked at a separate regression model, gradient boosted ensembles of regression trees, in order to benchmark vs. LR. For further reading, insightful discussion, and definitions for tree-based learning methods please see \cite{books/wa/BreimanFOS84}. 

%Performance criteria for our trained models included the $R^2$ score and Mean squared error (MSE) \cite{Free:2005} of each model tested on a held out test set using a 70/30 percentage train/test split.
%The $R^2$ score, also known as the coefficient of determination, gives an indication of model fit. It is asymmetric and ranges  $R^2 \in [-\infty$:1]. In order to calculate $R^2$, we first take the mean from our test vector $\mathbf{y}$.
%\begin{equation}
%    \bar{y} = \frac{1}{N}\sum_{i=1}^{n}{y_i}
%\end{equation}
%We may then calculate the total sum of squares,
%\begin{equation}
%    TSS =  \sum_{i}(y_i - \bar{y})^2
%\end{equation}
%And the residual sum of squares:
%\begin{equation}
%    RSS = \sum_{i}(y_i - \hat{y}_i)^2
%\end{equation}
%Where $\hat{y}_i$ is the prediction from our given model.
%$R^2$ is then:
%\begin{equation}
%    R^2 = 1 - \frac{RSS}{TSS}
%\end{equation}
%For MSE we simply take the the mean of the sum of the square error of the true values from the test $\mathbf{y}$ and the prediction $\mathbf{\hat{y}}$:
%\begin{equation}
%MSE = \frac{1}{N}RSS
%\end{equation}
Performance criteria for our trained models included the mean absolute error (MAE). Model validation was performed using cross validation on a randomized held out set of 3 folds.

In examining the Friedman 1 data set, we designed a performance metric in order to test whether or not the FS algorithm selected the first 5 features in the set, which we knew to be the optimal subset. We call this performance metric $\textit{Subset accuracy}$ which we define
using a $\textit{hit score}$ where we take the cardinality of the intersection of the set of the index of selected features $\mathbf{k_{selected}}$ and set of index of optimal features  $\mathbf{k_{opt}}$ divided by the cardinality of the optimal feature set.
\begin{equation}
hit score= \frac{card(\mathbf{k_{selected}} \cap \mathbf{k_{opt}})}{card(\mathbf{k_{opt}})}
\end{equation}
We then calculate a \textit{length score} by taking the absolute value of the cardinality of $\mathbf{k_{selected}}$ minus the cardinality of $\mathbf{k_{opt}}$ and subtracting this value from 1. 
\begin{equation}
length score = 1 - \frac{|card(\mathbf{k_{selected}}) - card(\mathbf{k_{opt}})|}{card(\mathbf{k_{opt}})} 
\end{equation}
We then simply sum the two and divide by two to obtain $\textit{Subset accuracy}$:
\begin{equation}
\textit{Subset accuracy} = \frac{(hitscore + length score)}{2}
\end{equation}

\subsubsection{Baselines, filter and wrapper methods}
For each quantum assisted feature selection method, we bootstrapped each run with 10 result sets, and for each result set queried the quantum processing unit (QPU) for 10000 shots. We evaluated each result set, and took the best overall result, which we averaged over each fold of cross validation. For each data set, we tested the quantum assisted feature selection methods and compared versus the following methods, each of which we cross-validated using 3 folds with replacement:
\begin{itemize}
\item{All features (All):}

We initially fit the estimator over all features in the training set of $\mathbf{X}$ and $\mathbf{y}$ in order to understand the test error and establish a baseline for performance criteria.

\item{Greedy Ranked Method (GR):}

We devised a simple ranking algorithm, where for  each feature column vector $\mathbf{x_k} \in \mathbf{X_k}$, we calculated the $MIC(\mathbf{x_k, y})$, and then sorted the features based on this relevancy criterion and selected the top $k$, in this case the top fifty percent of ranked features. We used this heuristic since we did not have an intuition as to what the best $k$ would be a priori.

\item{Recursive Feature Elimination( RFE):}
We also tested a wrapper style feature selection algorithm, in this case Recursive Feature Elimination (RFE) as shown in \cite{10.1023/A:1012487302797}. We did so in order to benchmark our quantum assisted method with the RFE wrapper method. For further implementation details please see \cite{10.1023/A:1012487302797}.
\end{itemize}

\subsection{Parameters for Experimentation}
For each data set and each estimator we manually tuned the parameters $\alpha$ and $\lambda$
within the QUBO formulation to values of 1000 and 10. With this we were able to obtain reasonable results over our baseline methods.This can be explained by the parameter $\alpha =1000$ producing a scaling effect, creating a more rugged objective landscape to optimize over. The parameter $\lambda=10$ had the effect of constraining the result sets to smaller or larger ranges for sizes of $k$. An interesting detail of experimentation resulted when testing various settings for $\lambda$; scaling to different values sometimes led to finding values for $k$ which were improvements in regards to model fit and predictive performance over restricting output to a pre-defined $k$. While we did not include optimizing these hyper-parameters in the scope of this work, we believe that future work could entail investigating this point in further detail. 

\subsection{Implementation Details}
Python Code for experiments was generated using the libraries scikit-learn \cite{scikit-learn} , pandas \cite{mckinney-proc-scipy-2010}, and scipy \cite{2020SciPy-NMeth}. Access to the Dwave quantum annealing machine was obtained using Leap software and API and the dimod library. All experiments were performed using the Dwave Advantage 1.1 sampler. Some statistical measures were calculated using the minepy library \cite{10.1093/gigascience/giy032}.

\section{Results}

For our experiments, we looked at performance metrics in terms of prediction error ($MAE$) and feature selection subset size $k$ over cross-validated held out test data for each of the two data sets. For the quantum-assisted routines, we took the average over the cross validation folds of the best sample from a  bootstrapped sample set of 10 samples. 

For the feature selection run times, we only considered the anneal time on the QPU in microseconds, versus wall clock time for the RFE and GR routines. We did not include statistics for all features, as there were no feature selection routines involved. Results showed that the quantum annealing times were constant with respect to problem size, at roughly 2000 microseconds of annealing times for all sampling runs vs the greedy and RFE routines, which varied with the size of the input and number of wrapper function evaluations in the range of 35000-50000 microseconds measured in wall-clock time for processing calls using the central processing unit (CPU) . This contrasts with the scaling by the quantum regime with respect to the problem size. Within the quantum assisted regime the space is bounded by the number of qubits available to encode the problem on the QPU, while the time is bound by the number of shots per sampling run, and anneal time per shot. 

Results for the Friedman 1 data set (in Table I) and UCI Auto data set (in Table II) show optimal results obtained using PCC as the correlation statistic within the QUBO formulation for the quantum-assisted routine. This is reasonable, since the underlying model is a regression algorithm, which implies that the mapping of inputs to the target variable is linear given the data sets. 

For the Friedman 1 data set, the quantum assisted routine using Pearson correlation coefficient within the QUBO formulation and multiple linear regression as the learning algorithm (\textbf{QPCC-LR})  achieved best performance with the lowest averaged Mean Absolute Error, $\mathit{MAE}$=2.27, and optimal size for $k=5$, with a Subset Accuracy score of 0.9.  For the UCI Auto data set the best score was obtained by the quantum assisted routine using Pearson correlation coefficient within the QUBO formulation and using the gradient boosted regression trees as a learning algorithm (\textbf{QPCC-GBR}) achieving the lowest $\mathit{MAE}$=1471 and size for $k$=5 Although the other distance metrics did not outperform PCC, some were comparable, as in the case with MIC in application to LR for the Friedman Data set, or MI for GBR on the Auto Data set. This hints that these metrics may be as robust as PCC for certain types of applications and data sets, and it may be that there are other data sets with which these statistics outperform PCC and emphasize the \textit{minimum redundancy} component of the MRMR heuristic. Overall, the quantum-assisted feature selection methods achieved results that outperformed our baselines of all features, greedy selection, and recursive feature selection.

%\begin{wraptable}{}{}
\begin{table}{}{}
\normalsize
\captionsetup{font=footnotesize}
\begin{tabular}{|p{2.2cm}||p{1.8cm}|p{1cm}|p{1cm}|}
\hline
\multicolumn{1}{|c|}{\textbf{Friedman 1 Data Set}}&&&\\
\hline 
\textbf{FS method}& \textit{MAE} & \textbf{$k$} & \textit{SA}\\
\hline
\hline
QMI-LR   &$2.58 \pm{0.12}$  & 5   & 0.9 \\
QMIC-LR   & $2.42 \pm{0.22}$ &  5  & 0.8 \\
QGMIC-LR   & $2.63 \pm{0.11}$  & 5   & 0.9 \\
\textbf{QPCC-LR}&$2.27 \pm{0.12}$  & 5    & 0.9 \\
All-LR &$3.461 \pm{0.44}$  & 26 &  - \\
GR-LR   &$2.73 \pm{0.19}$ & 12   & 0.4 \\
RFE-LR   &$3.22 \pm{0.40}$  & 12   & 0.4 \\
\hline
QMI-GBR   &$3.24 \pm{0.39}$  & 5   & 0.59 \\
QMIC-GBR &$2.92\pm{0.55}$  & 5   & 0.8 \\
QGMIC-GBR  &$2.99\pm{0.26}$  &5    & 0.8\\
QPCC-GBR &$2.42\pm{0.30}$   & 5 & 0.9    \\
All-GBR  &$2.96\pm{0.42}$   &26    & - \\
GR-GBR &$2.93\pm{0.52}$ & 12   & 0.4\\
RFE-GBR &$2.98\pm{0.39}$ & 12   & 0.4\\
\hline
\end{tabular}
\small
\caption{Table of Results for Friedman 1 Data Set 
%Feature selection method abbreviations follow the convention of "FS method/benchmark - learning algorithm" so in the case of All-LR, this would be all features with no feature selection, mapped to the output of a multiple linear regression model. For more details on acronym representation please see the Nomenclature section in the Appendix. We include the Subset Accuracy metric (SA) here to understand the ratio of how many optimal features were selected out of the optimal subset $\mathbf{k_{opt}}$. For this experiment, in the figure above, we see all of the quantum-assisted routines achieving approximately similar performance, with slight gains for QPPC-LR (Quantum-Assisted Pearson correlation coefficient- Multiple Linear Regression) with the lowest averaged $\mathbf{MAE = 2.27}$ and optimal size for $k= 5$, with a Subset Accuracy score of 0.9. 
}
\end{table}

\begin{table}{}{}
\normalsize
\captionsetup{font=footnotesize}
\begin{tabular}{|p{2.2cm}||p{2cm}|p{2cm}|}
\hline
\multicolumn{1}{|c|}{\textbf{UCI Auto Data Set}}&&\\
\hline
$\textbf{FS method}$& $\textbf{MAE}$ & $\textbf{k}$\\
\hline
\hline
QMI-LR   & $2669.1 \pm{417}$    &19\\
QMIC-LR & $2684.8 \pm{484}$&20\\
QGMIC-LR &$2669 \pm{482}$&20\\
QPCC-LR & $2622.3 \pm{463}$&18\\
All-LR &$2690.5 \pm{460}$&26\\
GR-LR& $2980.3 \pm{445}$&12\\
RFE-LR &$3355.7 \pm{527}$&12\\
\hline
\textbf{QPCC-GBR}  &$1471\pm{135.6}$&18\\
QMI-GBR   & $1491.5 \pm{142}$   &18\\
QGMIC-GBR &   $1491.5 \pm{181}$ & 19\\
QMIC-GBR   & $1536.2 \pm{168}$ & 20\\
All-GBR & $1546.2 \pm{154}$ & 26\\
GR-GBR&  $1707.2 \pm{168}$ & 12\\
RFE-GBR & $1678.3 \pm{143}$ & 12 \\
\hline

\end{tabular}
\small
\caption{Table of Results for UCI Auto Data set 
%Abbreviated FS methods are denoted by "FS method/benchmark - learning algorithm" for example QGMIC-GBR represents Quantum-Assisted Generalized Mean Information Coefficient- Multiple Linear Regression, meaning that the GMIC distance criterion was used to construct the QUBO for the quantum-assisted routine, which was fed into the gradient boosted regression tree learning algorithm. For more details on acronym representation please see the Nomenclature section in the Appendix. For this experiment, in the table above, we see the best score for the quantum assisted routine using Pearson correlation coefficient, with a gradient boosted decision tree learning algorithm obtaining the lowest $\mathit{MAE}$ =1471 and size for $k$=5.
}

\end{table}

\section{Conclusion}
In conclusion, we show that by leveraging quantum assisted routines within machine learning training and testing regimes, we achieve solutions which beat our defined benchmarks. We also uncovered that quantum-assisted routines may show the additional benefit of discovering sizes of $k$, given some slight tuning of $\alpha$ and $\lambda$, which show performance improvements. Future work could investigate optimizing these hyper-parameters using Bayesian optimization or other methodology in order to discover optimal subset sizes of $k$ automatically.
While we limited the scope of this work to focus on using the quantum annealer as the device for our quantum assisted routine, this problem could be formulated to run as an input problem Hamiltonian for a variational quantum algorithm on a gate model chip. Further work could also explore this point in more detail.

%Overall, we show the benefits of leveraging near-term quantum devices to assist classical machine learning regimes in application to real world problems such as price prediction. 

% or


\section{Appendix}
\makenomenclature
\nomenclature{\textbf{QMIR-LR}}{Quantum-assisted method leveraging  mutual information as distance metric and Multiple linear Regression as supervised learning algorithm}
\nomenclature{\textbf{QMIC-LR}}{Quantum-assisted method leveraging  maximal information coefficient as distance metric and Multiple linear Regression as supervised learning algorithm}
\nomenclature{\textbf{QGMIC-LR}}{Quantum-assisted method leveraging  generalized mean information coefficient as distance metric and Multiple linear Regression as supervised learning algorithm}
\nomenclature{\textbf{QPCC-LR}}{Quantum-assisted method leveraging  Pearson correlation coefficient as distance metric and Multiple linear Regression as supervised learning algorithm}
\nomenclature{\textbf{All-LR}}{All features with Multiple linear Regression as supervised learning algorithm}
\nomenclature{\textbf{RFE-LR}}{Recursive feature elimination as feature selection method with Multiple linear Regression as supervised learning algorithm}
\nomenclature{\textbf{GR-LR}}{Greedy feature selection method with Multiple linear Regression as supervised learning algorithm}
\nomenclature{\textbf{QMIR-GBR}}{Quantum-assisted method leveraging  mutual information as distance metric and Gradient boosted regression trees as supervised learning algorithm}
\nomenclature{\textbf{QMIC-GBR}}{Quantum-assisted method leveraging  maximal information coefficient as distance metric and Gradient boosted regression trees as supervised learning algorithm}
\nomenclature{\textbf{QGMIC-GBR}}{Quantum-assisted method leveraging  generalized mean information coefficient as distance metric and Gradient boosted regression trees as supervised learning algorithm}
\nomenclature{\textbf{QPCC-GBR}}{Quantum-assisted method leveraging  Pearson correlation coefficient as distance metric and Gradient boosted regression trees as supervised learning algorithm}
\nomenclature{\textbf{All-GBR}}{All features with Gradient boosted regression trees as supervised learning algorithm}
\nomenclature{\textbf{RFE-GBR}}{Recursive feature elimination as feature selection method with Gradient boosted regression trees as supervised learning algorithm}
\nomenclature{\textbf{GR-GBR}}{Greedy feature selection method with Gradient boosted regression trees as supervised learning algorithm}
\printnomenclature


\section*{Acknowledgment}
David Von Dollen would like to thank Abdallah Shanti, Tom Bartol, and Andre Radon for supporting this research and development effort, Reuben Brasher and Sheir Yarkoni for insightful discussion around the research topics.

\ifCLASSOPTIONcaptionsoff
  \newpage
\fi


%\addbibresource{biblio.bib}
\bibliographystyle{IEEEtran}
%\bibliography{IEEEtran,./bib.bbl}
%\printbibliography
%\begin{thebibliography}{1}


%\end{thebibliography}

%\bibliographystyle{unsrt}
{\footnotesize{\bibliography{biblio}}}

%\begin{biography}[{\includegraphics[width=1in,height=1.25in,clip,keepaspectratio]{mshell}}]{Michael Shell}

%\end{IEEEbiography}


%\newpage



%\vfill


\end{document}